\documentclass[10pt,twocolumn,letterpaper]{article}

\usepackage{iccv}
\usepackage{times}
\usepackage{epsfig}
\usepackage{graphicx}
\usepackage{amsmath}
\usepackage{amssymb}

\usepackage{booktabs}
\usepackage{multirow}

\usepackage[pagebackref=true,breaklinks=true,letterpaper=true,colorlinks,bookmarks=false]{hyperref}
\hypersetup{linkcolor=black}

\iccvfinalcopy 

\newcommand{\figLabel}{Figure\xspace}
\newcommand{\eqLabel}{Equation\xspace}
\newcommand{\secLabel}{Section\xspace}
\newcommand{\tblLabel}{Table\xspace}
\newcommand{\mysection}[1]{\vspace{3pt}\noindent\textbf{#1.}}

\newcommand{\res}{$\oplus$}
\newcommand{\dense}{$\rhd$\hspace{-2pt}$\lhd$}
\newcommand{\nc}{}

\def\rot#1{\rotatebox{0}{#1}}

\begin{document}

\title{DeepGCNs: Can GCNs Go as Deep as CNNs?\\\small\url{ https://sites.google.com/view/deep-gcns}}

\author{Guohao Li\thanks{equal contribution}\quad Matthias M\"uller\footnotemark[1]\quad Ali Thabet\quad Bernard Ghanem\\
		Visual Computing Center,~ KAUST,~ Thuwal,~ Saudi Arabia\\
		{\tt\footnotesize \{guohao.li, matthias.mueller.2, ali.thabet, bernard.ghanem\}@kaust.edu.sa}}

\maketitle

\begin{abstract}
Convolutional Neural Networks (CNNs) achieve impressive performance in a wide variety of fields. Their success benefited from a massive boost when very deep CNN models were able to be reliably trained. Despite their merits, CNNs fail to properly address problems with non-Euclidean data. To overcome this challenge, Graph Convolutional Networks (GCNs) build graphs to represent non-Euclidean data, borrow concepts from CNNs, and apply them in training. GCNs show promising results, but they are usually limited to very shallow models due to the vanishing gradient problem (see \figLabel \ref{fig:intro_fig}). As a result, most state-of-the-art GCN models are no deeper than $3$ or $4$ layers. In this work, we present new ways to successfully train very deep GCNs. We do this by borrowing concepts from CNNs, specifically residual/dense connections and dilated convolutions, and adapting them to GCN architectures. Extensive experiments  show the positive effect of these deep GCN frameworks. Finally, we use these new concepts to build a very deep 56-layer GCN, and show how it significantly boosts performance ($+3.7\%$ mIoU over state-of-the-art) in the task of point cloud semantic segmentation. We believe that the community can greatly benefit from this work, as it opens up many opportunities for advancing GCN-based research.
\end{abstract}

\section{Introduction}
\label{sec:introduction}
GCNs have been gaining a lot of momentum in the last few years. This increased interest is attributed to two main factors: the increasing proliferation of non-Euclidean data in real-world applications, and the limited performance of CNNs when dealing with such data. GCNs operate directly on non-Euclidean data and are very promising for applications that depend on this information modality. GCNs are currently used to predict individual relations in social networks \cite{social_tang2009relational}, model proteins for drug discovery \cite{chem_zitnik2017predicting,chem_wale2008comparison}, enhance predictions of recommendation engines \cite{rec_monti2017geometric,rec_ying2018graph}, efficiently segment large point clouds \cite{wang2018dynamic}, among other fields.

\begin{figure}[!t]
    \centering
    \begin{tabular}{cc}
    \includegraphics[trim=5mm 0mm 0mm 0mm, width=1\columnwidth]{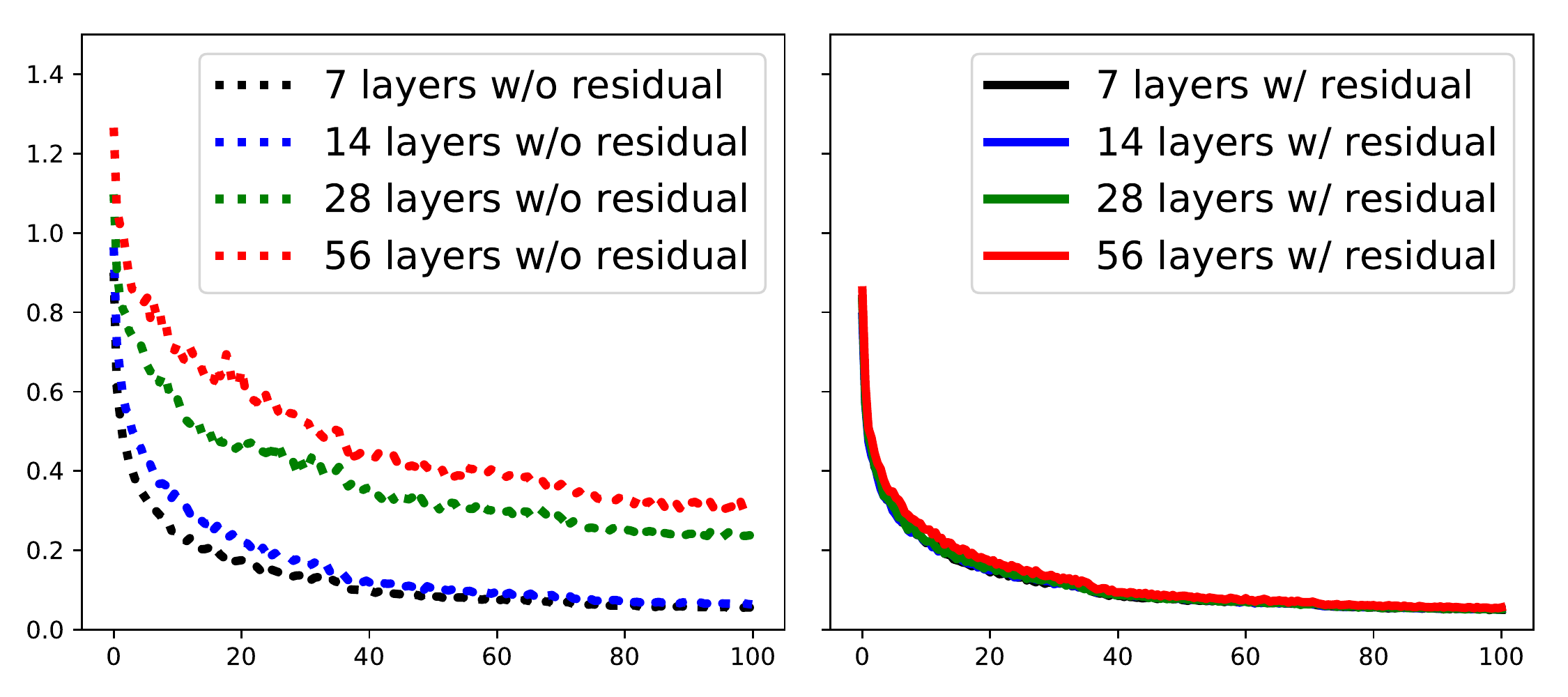}
    \end{tabular}
    \caption{\textbf{Training Deep GCNs}. (\emph{left}) We show the training loss for GCNs with 7, 14, 28, and 56 layers, with and without residual connections. We note how adding more layers without residual connections translates to substantially higher loss. (\emph{right}) In contrast, training GCNs with residual connections results in consistent stability across all depths.}
    \label{fig:intro_fig}
\end{figure}

A key reason behind the success of CNNs is the ability to design and reliably train very deep CNN models. In contrast, it is not yet clear how to properly train deep GCN architectures, where several works have studied their limitations  \cite{li2018deeper,wu2019comprehensive,zhou2018graph}. Stacking more layers into a GCN leads to the common vanishing gradient problem. This means that back-propagating through these networks causes over-smoothing, eventually leading to features of graph vertices converging to the same value \cite{li2018deeper}. Due to these limitations, most state-of-the-art GCNs are no deeper than $4$ layers \cite{zhou2018graph}.

Vanishing gradients is not a foreign phenomenon in the world of CNNs. It also posed limitations on the depth growth of these types of networks. ResNet \cite{he2016deep} provided a big step forward in the pursuit of very deep CNNs when it introduced residual connections between input and output layers. These connections massively alleviated the vanishing gradient problem. Today, ResNets can reach 152 layers and beyond. Further extension came with DenseNet \cite{huang2017densely}, where more connections are introduced across layers. More layers could potentially mean more spatial information loss due to pooling. This issue was also addressed, with Dilated Convolutions \cite{yu2015multi}. The introductions of these key concepts had substantial impact on the progress of CNNs, and we believe they can have a similar effect if well adapted to GCNs.

In this work, we present an extensive study of methodologies that allow for training very deep GCNs. We adapt concepts that were successful in training deep CNNs, mainly residual connections, dense connections, and dilated convolutions. We show how we can incorporate these layers into a graph framework, and present an extensive analysis of the effect of these additions to the accuracy and stability of deep GCNs. To showcase these layer adaptations, we apply them to the popular task of point cloud semantic segmentation. We show that adding a combination of residual and dense connections, and dilated convolutions, enables successful training of GCNs up to $56$ layers deep (refer to \figLabel \ref{fig:intro_fig}). This very deep GCN improves the state-of-the-art on the challenging S3DIS \cite{2017arXiv170201105A} point cloud dataset by $3.7\%$. 

\vspace{6pt}\mysection{Contributions} We summarize our contributions as three fold. \textbf{(1)} We adapt residual/dense connections, and dilated convolutions to GCNs. \textbf{(2)} We present extensive experiments on point cloud data, showing the effect of each of these new layers to the stability and performance of training deep GCNs. We use point cloud semantic segmentation as our experimental testbed. \textbf{(3)} We show how these new concepts help build a 56-layer GCN, the deepest GCN architecture by a large margin, and achieve close to $4\%$ boost in state-of-the-art performance on the S3DIS dataset \cite{2017arXiv170201105A}.

\section{Related Work}
\label{sec:related}
A large number of real-world applications deal with non-Euclidean data, which cannot be systematically and reliably processed by CNNs in general. To overcome the shortcomings of CNNs, GCNs provide well-suited solutions for non-Euclidean data processing, leading to greatly increasing interest in using GCNs for a variety of applications. In social networks \cite{social_tang2009relational}, graphs represent connections between individuals based on mutual interests/relations. These connections are non-Euclidean and highly irregular. GCNs help better estimate edge strengths between the vertices of social network graphs, thus leading to more accurate connections between individuals. Graphs are also used to model chemical molecule structures \cite{chem_zitnik2017predicting,chem_wale2008comparison}. Understanding the bio-activities of these molecules can have substantial impact on drug discovery. Another popular use of graphs is in recommendation engines \cite{rec_monti2017geometric,rec_ying2018graph}, where accurate modelling of user interactions leads to improved product recommendations. Graphs are also popular modes of representation in natural language processing \cite{nlp_bastings2017graph,nlp_marcheggiani2017encoding}, where they are used to represent complex relations between large text units. 

GCNs also find many applications in computer vision. In scene graph generation, semantic relations between objects are modelled using a graph. This graph is used to detect and segment objects in images, and also to predict semantic relations between object pairs \cite{qi20173d,cv_scene_xu2017scene,cv_scene_yang2018graph,cv_scene_li2018factorizable}. Scene graphs also facilitate the inverse process, where an image is reconstructed given a graph representation of the scene \cite{cv_inv_scene_johnson2018image}. Graphs are also used to model human joints for action recognition in video \cite{cv_action_yan2018spatial,cv_action_jain2016structural}.  
GCNs are a perfect candidate for 3D point cloud processing, especially since the unstructured nature of point clouds poses a representational challenge for systematic research. Several attempts in creating structure from 3D data exist by either representing it with multiple 2D views \cite{mv_su2015multi,mv_guerry2017snapnet,mv_boulch2017unstructured,mv_li2016lstm}, or by voxelization \cite{voxel_dai2017scannet,voxel_mv_qi2016volumetric,voxel_riegler2017octnet,voxel_tchapmi2017segcloud}. More recent work focuses on directly processing unordered point cloud representations \cite{pc_qi2017pointnet,pc_qi2017pointnet++, 3dsemseg_ICCVW17,pc_huang2018recurrent,pc_ye20183d}. The recent \emph{EdgeConv} method by Wang \etal \cite{wang2018dynamic} applies GCNs to point clouds. In particular, they propose a dynamic edge convolution algorithm for semantic segmentation of point clouds. The algorithm dynamically computes node adjacency at each graph layer using the distance between point features. This work demonstrates the potential of GCNs for point cloud related applications and beats the state-of-the-art in the task of point cloud segmentation. Unlike most other works, \emph{EdgeConv} does not rely on RNNs or complex point aggregation methods. 

Current GCN algorithms including \emph{EdgeConv} are limited to shallow depths. Recent works attempt to train deeper GCNs. For instance, Kipf \etal trained a semi-supervised GCN model for node classification and showed how performance degrades when using more than 3 layers \cite{kipf2016semi}. Pham \etal \cite{pham2017column} proposed Column Network (CLN) for collective classification in relational learning and showed peak performance with 10 layers with the performance degrading for deeper graphs. Rahimi \etal \cite{rahimi2018semi} developed a Highway GCN for user geo-location in social media graphs, where they add ``highway" gates between layers to facilitate gradient flow. Even with these gates, the authors demonstrate performance degradation after 6 layers of depth. Xu \etal \cite{xu2018representation} developed a \emph{Jump Knowledge Network} for representation learning and devised an alternative strategy to select graph neighbors for each node based on graph structure. As with other works, their network is limited to a small number of layers ($6$). Recently, Li \etal \cite{li2018deeper} studied the depth limitations of GCNs and showed that deep GCNs can cause over-smoothing, which results in features at vertices within each connected component converging to the same value. Other works \cite{wu2019comprehensive,zhou2018graph} also show the limitations of stacking multiple GCN layers, which lead to highly complex back-propagation and the common vanishing gradient problem. 

Many difficulties facing GCNs nowadays (\eg vanishing gradients and limited receptive field) were also present in the early days of CNNs \cite{he2016deep,yu2015multi}. We bridge this gap and show that the majority of these drawbacks can be remedied by borrowing several orthogonal tricks from CNNs. Deep CNNs achieved a huge boost in performance with the introduction of ResNet \cite{he2016deep}. By adding residual connections between inputs and outputs of layers, ResNet tends to alleviate the vanishing gradient problem. DenseNet \cite{huang2017densely} takes this idea a step further and adds connections across layers as well. Dilated Convolutions \cite{yu2015multi} are a more recent approach that has lead to significant performance gains, specifically in image-to-image translation tasks such as semantic segmentation \cite{yu2015multi}, by increasing the receptive field without loss of resolution. In this work, we show how one can benefit from concepts introduced for CNNs, mainly residual/dense connections and dilated convolutions, to train very deep GCNs. We support our claim by extending the work of Wang \etal \cite{wang2018dynamic} to a much deeper GCN, and therefore significantly increasing its performance. Extensive experiments on the task of point cloud semantic segmentation validate these ideas for general graph scenarios.

\section{Methodology} 
\label{sec:methodology}

\begin{figure*}
    \centering
    \includegraphics[page=2,trim = 10mm 10mm 40mm 10mm, clip, width=\textwidth]{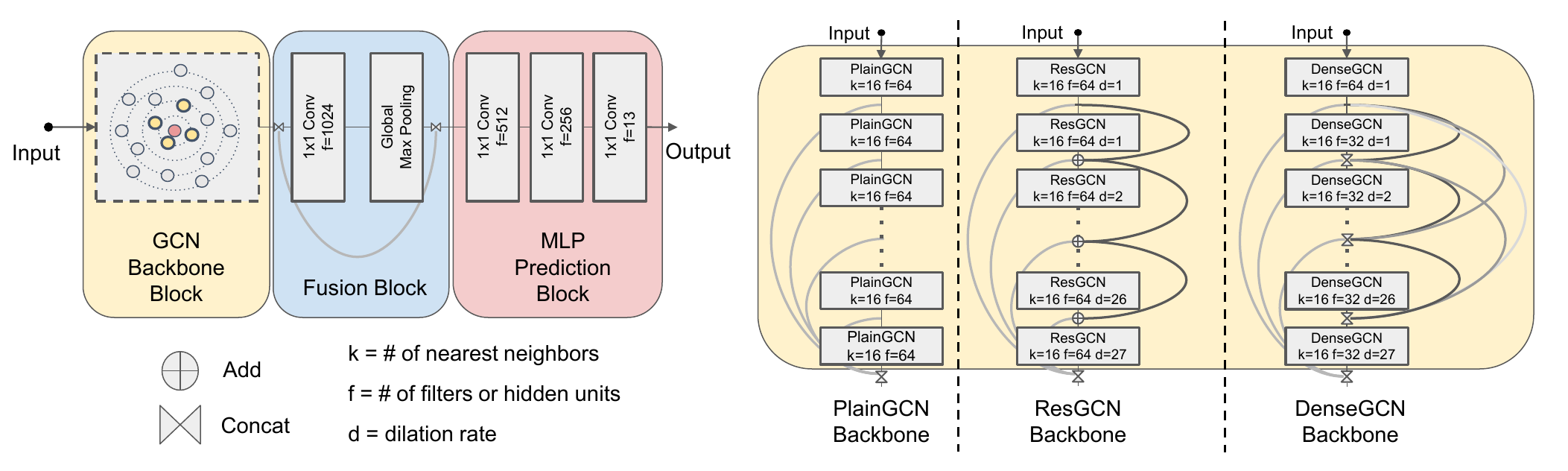}
    \caption{\textbf{Proposed GCN  architecture for point cloud semantic segmentation}. \textit{(left)} Our framework consists of three blocks: a GCN Backbone Block (feature transformation of input point cloud), a Fusion Block (global feature generation and fusion), and an MLP Prediction Block (point-wise label prediction). \textit{(right)} We study three types of GCN Backbone Block (\emph{PlainGCN}, \emph{ResGCN} and \emph{DenseGCN}) and use two kinds of layer connection (vertex-wise addition used in \emph{ResGCN} or vertex-wise concatenation used in \emph{DenseGCN}). 
    }
\label{fig:pipeline}
\end{figure*}

\subsection{Representation Learning on Graphs} \label{sec:defineGCN}
\mysection{Graph Definition} A graph $\mathcal{G}$ is represented by a tuple $\mathcal{G}=(\mathcal{V}, \mathcal{E})$ where $\mathcal{V}$ is the set of unordered vertices and $\mathcal{E}$ is the set of edges representing the connectivity between vertices $v \in \mathcal{V}$. If $e_{i,j} \in \mathcal{E}$, then vertices $v_i$ and $v_j$ are connected to each other with an edge $e_{i,j}$.

\mysection{Graph Convolution Networks} Inspired by CNNs, GCNs intend to extract richer features at a vertex by aggregating features of vertices from its neighborhood. GCNs represent vertices by associating each vertex $v$ with a feature vector $\mathbf{h}_{v} \in \mathbb{R}^{D}$,
where $D$ is the feature dimension. Therefore, the graph $\mathcal{G}$ as a whole can be represented by concatenating the features of all the unordered vertices, \ie $\mathbf{h}_{\mathcal{G}} = [\mathbf{h}_{v_1}, \mathbf{h}_{v_2}, ..., \mathbf{h}_{v_N}]^{\top} \in \mathbb{R}^{N\times D}$,
where $N$ is the cardinality of set $\mathcal{V}$. A general graph convolution operation $\mathcal{F}$ at the $l$-th layer can be formulated as the following aggregation and update operations,
\begin{equation} \label{eq:GCN}
\begin{split}
\mathcal{G}_{l+1} &=\mathcal{F}(\mathcal{G}_{l}, \mathcal{W}_l)\\
&= Update(Aggregate(\mathcal{G}_{l}, \mathcal{W}_l^{agg}), \mathcal{W}_l^{update}).
\end{split}
\end{equation}
$\mathcal{G}_l = (\mathcal{V}_{l}, \mathcal{E}_{l})$ and $\mathcal{G}_{l+1} = (\mathcal{V}_{l+1}, \mathcal{E}_{l+1})$ are the input and output graphs at the $l$-th layer, respectively.  $\mathcal{W}_l^{agg}$ and $\mathcal{W}_l^{update}$ are the learnable weights of the aggregation and update functions respectively, and they are the essential components of GCNs. In most GCN frameworks, aggregation functions are used to compile information from the neighborhood of vertices, while update functions perform a non-linear transform on the aggregated information to compute new vertex representations. There are different variants of those two functions. For example, the aggregation function can be a mean aggregator \cite{kipf2016semi}, a max-pooling aggregator \cite{pc_qi2017pointnet,hamilton2017inductive,wang2018dynamic}, an attention aggregator \cite{velivckovic2017graph} or an LSTM aggregator \cite{peng2017cross}. The update function can be a multi-layer perceptron \cite{hamilton2017inductive,duvenaud2015convolutional}, a gated network \cite{li2015gated}, \etc. More concretely, the representation of vertices is computed at each layer by aggregating features of neighbor vertices for all $v_{l+1} \in \mathcal{V}_{l+1} $ as follows,
\begin{equation} \label{GCN_node}
\resizebox{.89\columnwidth}{!} {$
\mathbf{h}_{v_{l+1}} =\phi\left(\mathbf{h}_{v_{l}}, \rho(\{\mathbf{h}_{u_{l}}| u_{l}\in \mathcal{N}(v_{l})\}, \mathbf{h}_{v_{l}}, \mathcal{W}_{\rho}), \mathcal{W}_{\phi}\right)
$},
\end{equation}
where $\rho$ is a vertex feature aggregation function and $\phi$ is a vertex feature update function, $\mathbf{h}_{v_{l}}$ and $\mathbf{h}_{v_{l+1}}$ are the vertex features at the $l$-th layer and $l+1$-th layer respectively. $\mathcal{N}(v_{l})$ is the set of neighbor vertices of $v$ at the $l$-th layer, and $\mathbf{h}_{u_l}$ is the feature of those neighbor vertices parametrized by $\mathcal{W}_{\rho}$. $\mathcal{W}_{\phi}$ contains the learnable parameters of these functions. For simplicity and without loss of generality, we use a max-pooling vertex feature aggregator, without learnable parameters, to pool the difference of features between vertex $v_l$ and all of its neighbors: $\rho(.)=\max(\mathbf{h}_{u_{l}} - \mathbf{h}_{v_{l}}|~ u_{l}\in \mathcal{N}(v_{l}))$. We then model the vertex feature updater $\phi$ as a multi-layer perceptron (MLP) with batch normalization \cite{ioffe2015batch} and a ReLU as an activation function. This MLP concatenates $\mathbf{h}_{v_{l}}$ with its aggregate features from $\rho(.)$ to form its input.

\mysection{Dynamic Edges} \label{dyna} As mentioned earlier, most GCNs have fixed graph structures and only update the vertex features at each iteration. Recent work \cite{simonovsky2017dynamic,wang2018dynamic,valsesia2018learning} demonstrates that dynamic graph convolution, where the graph structure is allowed to change in each layer, can learn better graph representations compared to GCNs with fixed graph structure. For instance, ECC (Edge-Conditioned
Convolution) \cite{simonovsky2017dynamic} uses dynamic edge-conditional filters to learn an edge-specific weight matrix. Moreover, EdgeConv \cite{wang2018dynamic} finds the nearest neighbors in the current feature space to reconstruct the graph after every EdgeConv layer. In order to learn to generate point clouds, Graph-Convolution GAN (Generative Adversarial Network) \cite{valsesia2018learning} also applies $k$-NN graphs to construct the neighbourhood of each vertex in every layer. We find that dynamically changing neighbors in GCNs helps alleviate the over-smoothing problem and results in an effectively larger receptive field, when deeper GCNs are considered. In our framework, we propose to re-compute edges between vertices via a \emph{Dilated $k$-NN} function in the feature space of each layer to further increase the receptive field. 
In what follows, we provide detailed description of three operations that can enable much deeper GCNs to be trained: residual connections, dense connections, and dilated aggregation.

\subsection{Residual Learning for GCNs} \label{sec:ResGCN}
Designing deep GCN architectures \cite{wu2019comprehensive, zhou2018graph} is an open problem in the graph learning space. Recent work \cite{li2018deeper, wu2019comprehensive, zhou2018graph} suggests that GCNs do not scale well to deep architectures, since stacking multiple layers of graph convolutions leads to high complexity in back-propagation. As such, most state-of-the-art GCN models are usually no more than 3 layers deep \cite{zhou2018graph}. Inspired by the huge success of ResNet \cite{he2016deep}, DenseNet \cite{huang2017densely} and Dilated Convolutions \cite{yu2015multi}, we transfer these ideas to GCNs to unleash their full potential. This enables much deeper GCNs that reliably converge in training and achieve superior performance in inference. 

In the original graph learning framework, the underlying mapping $\mathcal{F}$, which takes a graph as an input and outputs a new graph representation (see \eqLabel \eqref{eq:GCN}), is learned.
Here, we propose a graph residual learning framework that learns an underlying mapping $\mathcal{H}$ by fitting another mapping $\mathcal{F}$. After $\mathcal{G}_{l}$ is transformed by $\mathcal{F}$, vertex-wise addition is performed to obtain $\mathcal{G}_{l+1}$. The residual mapping $\mathcal{F}$ learns to take a graph  as input and outputs a residual graph representation $\mathcal{G}_{l+1}^{res}$ for the next layer. $\mathcal{W}_l$ is the set of learnable parameters at layer $l$. In our experiments, we refer to our residual model as \emph{ResGCN}.
\begin{equation} \label{eq2}
\begin{split}
\mathcal{G}_{l+1} &= \mathcal{H}(\mathcal{G}_{l}, \mathcal{W}_l) \\
&= \mathcal{F}(\mathcal{G}_{l}, \mathcal{W}_l) + \mathcal{G}_{l}=\mathcal{G}_{l+1}^{res}+ \mathcal{G}_{l}.
\end{split}
\end{equation}

\subsection{Dense Connections in GCNs} \label{sec:DenseGCN}
DenseNet \cite{huang2017densely} was proposed to exploit dense connectivity among layers, which improves information flow in the network and enables efficient reuse of features among layers. Inspired by DenseNet, we adapt a similar idea to GCNs so as to exploit information flow from different GCN layers. In particular, we have:
\begin{equation} \label{eq4}
\begin{split}
\mathcal{G}_{l+1} &= \mathcal{H}(\mathcal{G}_{l}, \mathcal{W}_l) \\
&= \mathcal{T}(\mathcal{F}(\mathcal{G}_{l}, \mathcal{W}_l), \mathcal{G}_{l})  \\
&= \mathcal{T}(\mathcal{F}(\mathcal{G}_{l}, \mathcal{W}_l), ...,\mathcal{F}(\mathcal{G}_{0}, \mathcal{W}_{0}), \mathcal{G}_{0}).
\end{split}
\end{equation}
The operator $\mathcal{T}$ is a vertex-wise concatenation function that densely fuses the input graph $\mathcal{G}_0$ with all the intermediate GCN layer outputs. To this end, $\mathcal{G}_{l+1}$ consists of all the GCN transitions from previous layers. Since we fuse GCN representations densely, we refer to our dense model as \emph{DenseGCN}. The growth rate of \emph{DenseGCN} is equal to the dimension $D$ of the output graph (similar to DenseNet for CNNs \cite{huang2017densely}). For example, if $\mathcal{F}$ produces a $D$ dimensional vertex feature, where the vertices of the input graph $\mathcal{G}_0$ are $D_{0}$ dimensional, the dimension of each vertex feature of $\mathcal{G}_{l+1}$ is $D_{0}+D\times (l+1)$. 

\subsection{Dilated Aggregation in GCNs} \label{sec:dilation}

Dilated wavelet convolution is an algorithm originating from the wavelet processing domain \cite{holschneider1990real,shensa1992discrete}. To alleviate spatial information loss caused by pooling operations, Yu \etal \cite{yu2015multi} propose dilated convolutions as an alternative to applying consecutive pooling layers for dense prediction tasks, \eg semantic image segmentation. Their experiments demonstrate that aggregating multi-scale contextual information using dilated convolutions can significantly increase the accuracy of semantic segmentation tasks. The reason behind this is the fact that dilation enlarges the receptive field without loss of resolution. We believe that dilation can also help with the receptive fields of deep GCNs. Therefore, we introduce dilated aggregation to GCNs. There are many possible ways to construct a dilated neighborhood. We use a \emph{Dilated $k$-NN} to find dilated neighbors after every GCN layer and construct a \emph{Dilated Graph}. In particular, for an input graph $\mathcal{G}=(\mathcal{V}, \mathcal{E})$  with \emph{Dilated $k$-NN} and $d$ as the dilation rate, the \emph{Dilated $k$-NN} returns the $k$ nearest neighbors within the $k\times d$ neighborhood region by skipping every $d$ neighbors. The nearest neighbors are determined based on a pre-defined distance metric. In our experiments, we use the $\ell_2$ distance in the feature space of the current layer. 

\begin{figure}[!htb]
    \centering
    \includegraphics[page=2,trim = 25mm 10mm 30mm 1mm, clip, width=\columnwidth]{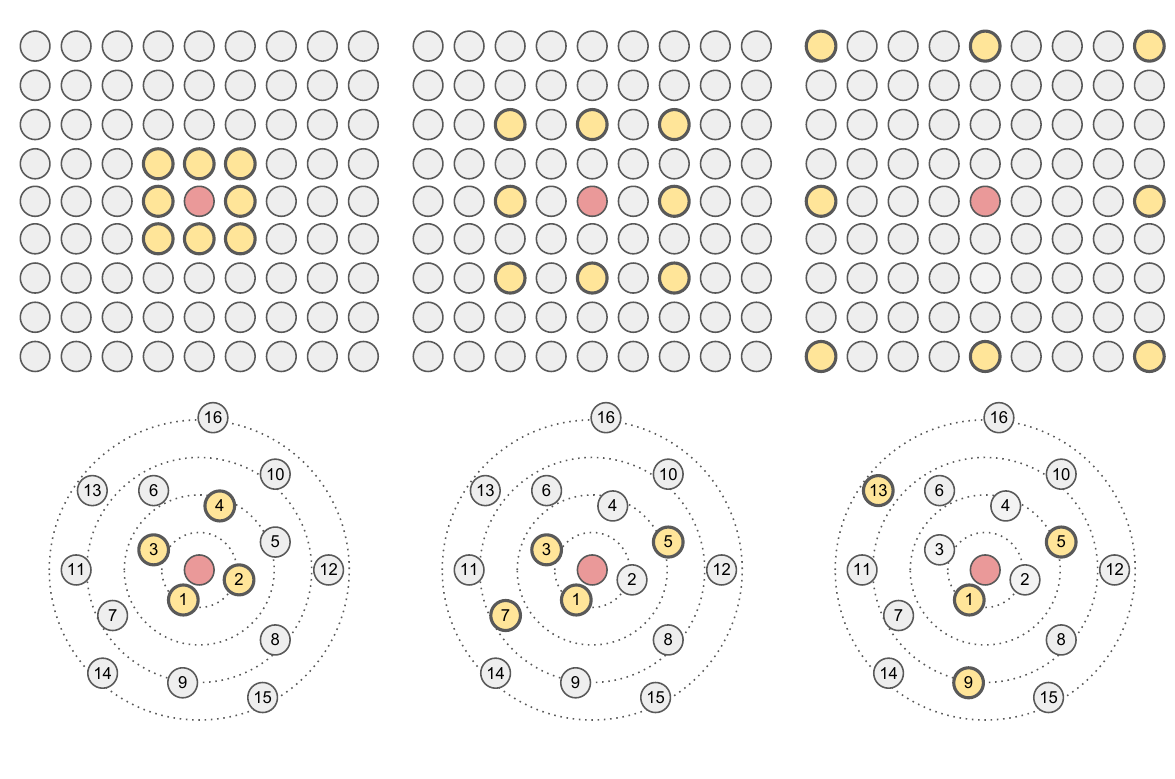}
    \caption{\textbf{Dilated Convolution in GCNs}. Visualization of dilated convolution on a structured graph arranged in a grid (\eg 2D image) and on a general structured graph. (\emph{top}) 2D convolution with kernel size 3 and dilation rate 1, 2, 4 (left to right). (\emph{bottom}) Dynamic graph convolution with dilation rate 1, 2, 4 (left to right).}
\label{fig:dilation_viz}
\end{figure}

Let $\mathcal{N}^{(d)}(v)$ denote the $d$-dilated neighborhood of vertex $v$. If $(u_1, u_2, ..., u_{k\times d})$ are the first sorted $k\times d$ nearest neighbors, vertices $(u_1, u_{1+d}, u_{1+2d}, ..., u_{1+(k-1)d})$ are the $d$-dilated neighbors of vertex $v$ (see \figLabel \ref{fig:dilation_viz}), \ie 
\begin{align*} \label{eq5}
    \mathcal{N}^{(d)}(v) = \{u_1, u_{1+d}, u_{1+2d}, ..., u_{1+(k-1)d}\}.
\end{align*}

Therefore, the edges $\mathcal{E}^{(d)}$ of the output graph are defined on the set of $d$-dilated vertex neighbors $\mathcal{N}^{(d)}(v)$. Specifically, there exists a directed edge $e \in \mathcal{E}^{(d)}$ from vertex $v$ to every vertex $u \in \mathcal{N}^{(d)}(v)$. The GCN aggregation and update functions are applied, as in \eqLabel \eqref{eq:GCN}, by using the edges $\mathcal{E}^{(d)}$ created by the \emph{Dilated $k$-NN}, so as to generate the feature $\mathbf{h}_{v}^{(d)}$ of each output vertex in $\mathcal{V}^{(d)}$. We denote this layer operation as a \emph{dilated graph convolution} with dilation rate $d$, or more formally: $\mathcal{G}^{(d)}=(\mathcal{V}^{(d)}, \mathcal{E}^{(d)})$. To improve generalization, we use \emph{stochastic dilation} in practice. During training, we perform the aforementioned dilated aggregations with a high probability $(1-\epsilon)$ leaving a small probability $\epsilon$ to perform random aggregation by uniformly sampling $k$ neighbors from the set of $k\times d$ neighbors $\{u_1, u_2, ..., u_{k\times d}\}$. At inference time, we perform deterministic dilated aggregation without stochasticity.

\section{Experiments}
\label{sec:experiments}

We propose \emph{ResGCN} and \emph{DenseGCN} to handle the vanishing gradient problem of GCNs. To enlarge the receptive field, we define a dilated graph convolution operator for GCNs. To evaluate our framework, we conduct extensive experiments on the task of large-scale point cloud segmentation and demonstrate that our methods significantly improve performance. In addition, we also perform a comprehensive ablation study to show the effect of different components of our framework.

\subsection{Graph Learning on 3D Point Clouds}
\label{sec:PCGCN}
Point cloud segmentation is a challenging task because of the unordered and irregular structure of 3D point clouds. Normally, each point in a point cloud is represented by its 3D spatial coordinates and possibly auxiliary features such as color and surface normal. We treat each point as a vertex $v$ in a directed graph $\mathcal{G}$ and we use $k$-NN to construct the directed dynamic edges between points at every GCN layer (refer to \secLabel \ref{dyna}). In the first layer, we construct the input graph $\mathcal{G}_{0}$ by executing a dilated $k$-NN search to find the nearest neighbor in 3D coordinate space. At subsequent layers, we dynamically build the edges using dilated $k$-NN in feature space. For the segmentation task, we predict the categories of all the vertices at the output layer. 

\subsection{Experimental Setup}
We use the overall accuracy (OA) and  mean intersection over union (mIoU) across all classes  as evaluation metrics. For each class, the IoU is computed as $\frac{TP}{TP+T-P}$, where $TP$ is the number of true positive points, $T$ is the number of ground truth points of that class, and $P$ is the number of predicted positive points. To motivate the use of deep GCNs, we do a thorough ablation study on area 5 to analyze each component and provide insights. We then evaluate our proposed reference model (backbone of 28 layers with residual graph connections and stochastic dilated graph convolutions) on all 6 areas and compare it to the shallow DGCNN baseline \cite{wang2018dynamic} and other state-of-the-art methods. 

\subsection{Network Architectures}
As shown in \figLabel \ref{fig:pipeline}, all the network architectures in our experiments have three blocks: a GCN backbone block, a fusion block and an MLP prediction block. The GCN backbone block is the only part that differs between experiments. For example, the only difference between \emph{PlainGCN} and \emph{ResGCN} is the use of residual skip connections for all GCN layers in \emph{ResGCN}. Both have the same number of parameters. We linearly increase the dilation rate $d$ of dilated $k$-NN with network depth. For fair comparison, we keep the fusion and MLP prediction blocks the same for all architectures. In the S3DIS semantic segmentation task, the GCN backbone block takes as input a point cloud with 4096 points, extracts features by applying consecutive GCN layers to aggregate local information, and outputs a learned graph representation with 4096 vertices. The fusion and MLP prediction blocks follow a similar architecture as PointNet \cite{pc_qi2017pointnet} and DGCNN \cite{wang2018dynamic}. The fusion block is used to fuse the global and multi-scale local features. It takes as input the extracted vertex features from the GCN backbone block at every GCN layer and concatenates those features, then passes them through a 1$\times$1 convolution layer followed by max pooling. The latter layer aggregates the vertex features of the whole graph into a single global feature vector, which in return is concatenated with the feature of each vertex from all previous GCN layers (fusion of global and local information). The MLP prediction block applies three MLP layers to the fused features of each vertex/point to predict its category. In practice, these layers are  1$\times$1 convolutions.

\mysection{PlainGCN} This baseline model consists of a \emph{PlainGCN} backbone block, a fusion block, and a MLP prediction block. 
The backbone stacks 28 EdgeConv \cite{wang2018dynamic} layers with dynamic $k$-NN, each of which is similar to the one used in DGCNN \cite{wang2018dynamic}. 
No skip connections are used here. 

\mysection{ResGCN} We construct \emph{ResGCN} by adding dynamic dilated $k$-NN and residual graph connections to \emph{PlainGCN}. These connections between all GCN layers in the GCN backbone block do not increase the number of parameters.  

\mysection{DenseGCN} Similarly, \emph{DenseGCN} is built by adding dynamic dilated $k$-NN and dense graph connections to the \emph{PlainGCN}. As described in \secLabel \ref{sec:DenseGCN}, dense graph connections are created by concatenating all the intermediate graph representations from previous layers. The dilation rate schedule of our \emph{DenseGCN} is the same as \emph{ResGCN}.

\subsection{Implementation}
We implement all our models using Tensorflow. For fair comparison, we use the Adam optimizer with the same initial learning rate $0.001$ and the same learning rate schedule; the learning rate decays $50\%$ every $3\times 10^5$ gradient decent steps. The networks are trained with two NVIDIA Tesla V100 GPUs using data parallelism. The batch size is set to $8$ for each GPU. Batch Normalization is applied to every layer. Dropout with a rate of $0.3$ is used at the second MLP layer of the MLP prediction block. As mentioned in \secLabel \ref{sec:dilation}, we use dilated $k$-NN with a random uniform sampling probability $\epsilon=0.2$ for GCNs with dilations. In order to isolate the effect of the proposed deep GCN architectures, we do not use any data augmentation or post processing techniques. We train our models end-to-end from scratch.

\begin{table*}[!htb]
\centering
\footnotesize 
\setlength{\tabcolsep}{7pt} 
\begin{tabular}{c|lcc|cccccccc}
\toprule
\textbf{\rot{Ablation}} & \textbf{\rot{Model}} & \textbf{\rot{mIoU}} & \textbf{\rot{$\Delta$mIoU}} & \textbf{\rot{dynamic}} & \textbf{\rot{connection}} & \textbf{\rot{dilation}} & \textbf{\rot{stochastic}} & \textbf{\rot{\# NNs}} & \textbf{\rot{\# filters}} & \textbf{\rot{\# layers}} \\
\midrule
\textbf{Reference}                 & \emph{ResGCN-28} & \textbf{52.49} & 0.00 & \checkmark  & \res & \checkmark & \checkmark & 16 & 64 & 28 \\
\midrule
\multirow{3}{*}{\textbf{Dilation}} &        & 51.98 & -0.51 & \checkmark  & \res & \checkmark &            & 16 & 64 & 28 \\
                                   &        & 49.64 & -2.85 & \checkmark  & \res &            &            & 16 & 64 & 28 \\ 
                                   & \emph{PlainGCN-28} & \textbf{40.31} & -12.18 & \checkmark  & \nc &            &            & 16 & 64 & 28 \\
\midrule
\multirow{2}{*}{\textbf{Fixed $k$-NN}}  & & 48.38 & -4.11 &             & \res &            &            & 16 & 64 & 28 \\   
                                          & & 43.43 & -9.06 &             & \nc     &            &            & 16 & 64 & 28 \\  
\midrule
\multirow{5}{*}{\textbf{Connections}} & \emph{DenseGCN-28} & \textbf{51.27} & -1.22 & \checkmark  & \dense    & \checkmark & \checkmark &  8 & 32 & 28 \\
                                      &     & 40.47 & -12.02 & \checkmark  & \nc     & \checkmark & \checkmark & 16 & 64 & 28 \\
                                      &     & 38.79 & -13.70 & \checkmark  & \nc     & \checkmark & \checkmark &  8 & 64 & 56 \\
                                      &     & 49.23 & -3.26 & \checkmark  & \nc     & \checkmark & \checkmark & 16 & 64 & 14 \\
                                      &     & 47.92 & -4.57 & \checkmark  & \nc     & \checkmark & \checkmark & 16 & 64 & 7 \\ 
\midrule
\multirow{2}{*}{\textbf{Neighbors}}&        & 49.98 & -2.51  & \checkmark & \res & \checkmark & \checkmark & 8 & 64 & 28 \\
                                   &        & 49.22 & -3.27  & \checkmark & \res & \checkmark & \checkmark & 4 & 64 & 28 \\
\midrule
\multirow{3}{*}{\textbf{Depth}} & \emph{ResGCN-56}          & \textbf{53.64} & 1.15 & \checkmark  & \res & \checkmark & \checkmark & 8 & 64 & 56 \\
                                & \emph{ResGCN-14}          & 49.90 & -2.59 & \checkmark  & \res & \checkmark & \checkmark & 16 & 64 & 14 \\
                                & \emph{ResGCN-7}          & 48.95 & -3.53 & \checkmark  & \res & \checkmark & \checkmark & 16 & 64 & 7 \\
\midrule
\multirow{4}{*}{\textbf{Width}} & \emph{ResGCN-28W} & \textbf{53.78} & 1.29 & \checkmark  & \res & \checkmark & \checkmark & 8 & 128 & 28 \\
                                &           & 49.18 & -3.31 & \checkmark  & \res & \checkmark & \checkmark & 32 & 32 & 28 \\
                                &           & 48.80 & -3.69 & \checkmark  & \res & \checkmark & \checkmark & 16 & 32 & 28 \\
                                &           & 45.62 & -6.87 & \checkmark  & \res & \checkmark & \checkmark & 16 & 16 & 28 \\
\bottomrule
\end{tabular}
\vspace{3pt}
\caption{\textbf{Ablation study on area 5 of S3DIS}. We compare our reference network (\emph{ResGCN-28}) with 28 layers, residual graph connections, and dilated graph convolutions to several ablated variants. All models were trained with the same hyper-parameters for 100 epochs on all areas except for area 5, which is used for evaluation. We denote residual and dense connections with the \res~ and \dense~ symbols respectively. We highlight the most important results in bold. $\Delta$mIoU denotes the difference in mIoU with respect to the reference model \emph{ResGCN-28}.}
\label{tbl:ablation_area5}
\end{table*}

\subsection{Results}
For convenient referencing, we use the naming convention \emph{BackboneBlock-\#Layers} to denote the key models in our analysis and we provide all names in \tblLabel \ref{tbl:ablation_area5}. We focus on residual graph connections for our analysis, since \emph{ResGCN-28} 
is easier and faster to train, but we expect that our observations also hold for dense graph connections.

We investigate the performance of different \emph{ResGCN} architectures, \eg with dynamic dilated $k$-NN, with regular dynamic $k$-NN (without dilation), and with fixed edges. We also study the effect of different parameters, \eg number of $k$-NN neighbors (4, 8, 16, 32), number of filters (32, 64, 128), and number of layers (7, 14, 28, 56). 
Overall, we conduct 20 experiments and show their results in \tblLabel \ref{tbl:ablation_area5}. 
\mysection{Effect of residual graph connections}
Our experiments in \tblLabel \ref{tbl:ablation_area5} (\textit{Reference})
show that residual graph connections play an essential role in training deeper networks, as they tend to result in more stable gradients. This is analogous to the insight from CNNs \cite{he2016deep}. When the residual graph connections between layers are removed (\ie in \emph{PlainGCN-28}), performance dramatically degrades (-12\% mIoU). In Appendices \ref{appendix:gcn_variants} and \ref{appendix:results_gcn_variants}, we  show similar performance gains by combining residual graph connections and dilated graph convolutions with other types of GCN layers.

\mysection{Effect of dilation}
Results in \tblLabel \ref{tbl:ablation_area5} (\textit{Dilation}) \cite{yu2015multi} show that dilated graph convolutions account for a 2.85\% improvement in mean IoU (\textit{row 3}), 
motivated primarily by the expansion of the network's receptive field. 
We find that adding stochasticity to the dilated $k$-NN does help performance but not to a significant extent. Interestingly, our results in \tblLabel \ref{tbl:ablation_area5} also indicate that dilation especially helps deep networks when combined with residual graph connections (\textit{rows 1,8}). Without such connections, performance can actually degrade with dilated graph convolutions. The reason for this is probably that these varying neighbors result in `worse' gradients, which further hinder convergence when residual graph connections are not used.

\mysection{Effect of dynamic $k$-NN}
While we observe an improvement when updating the $k$ nearest neighbors after every layer, we would also like to point out that it comes at a relatively high computational cost. We show different variants without dynamic edges in \tblLabel \ref{tbl:ablation_area5} (\textit{Fixed $k$-NN}).

\mysection{Effect of dense graph connections}
We observe similar performance gains with dense graph connections (\emph{DenseGCN-28}) in \tblLabel \ref{tbl:ablation_area5} (\textit{Connections}). However, with a naive implementation, the memory cost is prohibitive. Hence, the largest model we can fit into GPU memory uses only $32$ filters and $8$ nearest neighbors, as compared to $64$ filters and $16$ neighbors in the case of its residual counterpart \emph{ResGCN-28}. Since the performance of these two deep GCN variants is similar, residual connections are more practical for most use cases and, hence we focus on them in our ablation study. Yet, we do expect the same insights to transfer to the case of dense graph connections.

\mysection{Effect of nearest neighbors} Results in \tblLabel \ref{tbl:ablation_area5} (\textit{Neighbors}) show that  a larger number of neighbors helps in general. As the number of neighbors is decreased by a factor of 2 and 4, the performance drops by 2.5\% and 3.3\% respectively. However, a large number of neighbors only results in a performance boost, if the network capacity is sufficiently large. This becomes apparent when we increase the number of neighbors by a factor of 2 and decrease the number of filters by a factor of 2.

\mysection{Effect of network depth}
\tblLabel \ref{tbl:ablation_area5} (\textit{Depth}) shows that increasing the number of layers improves network performance, but only if residual graph connections and dilated graph convolutions are used, as in \tblLabel \ref{tbl:ablation_area5} (\textit{Connections}).

\mysection{Effect of network width} Results in \tblLabel \ref{tbl:ablation_area5}  (\textit{Width}) show that increasing the number of filters leads to a similar increase in performance as increasing the number of layers. In general, a higher network capacity enables learning nuances necessary for succeeding in corner cases.

\begin{figure*}[!h]
    \centering
    \includegraphics[page=1, width=\textwidth]{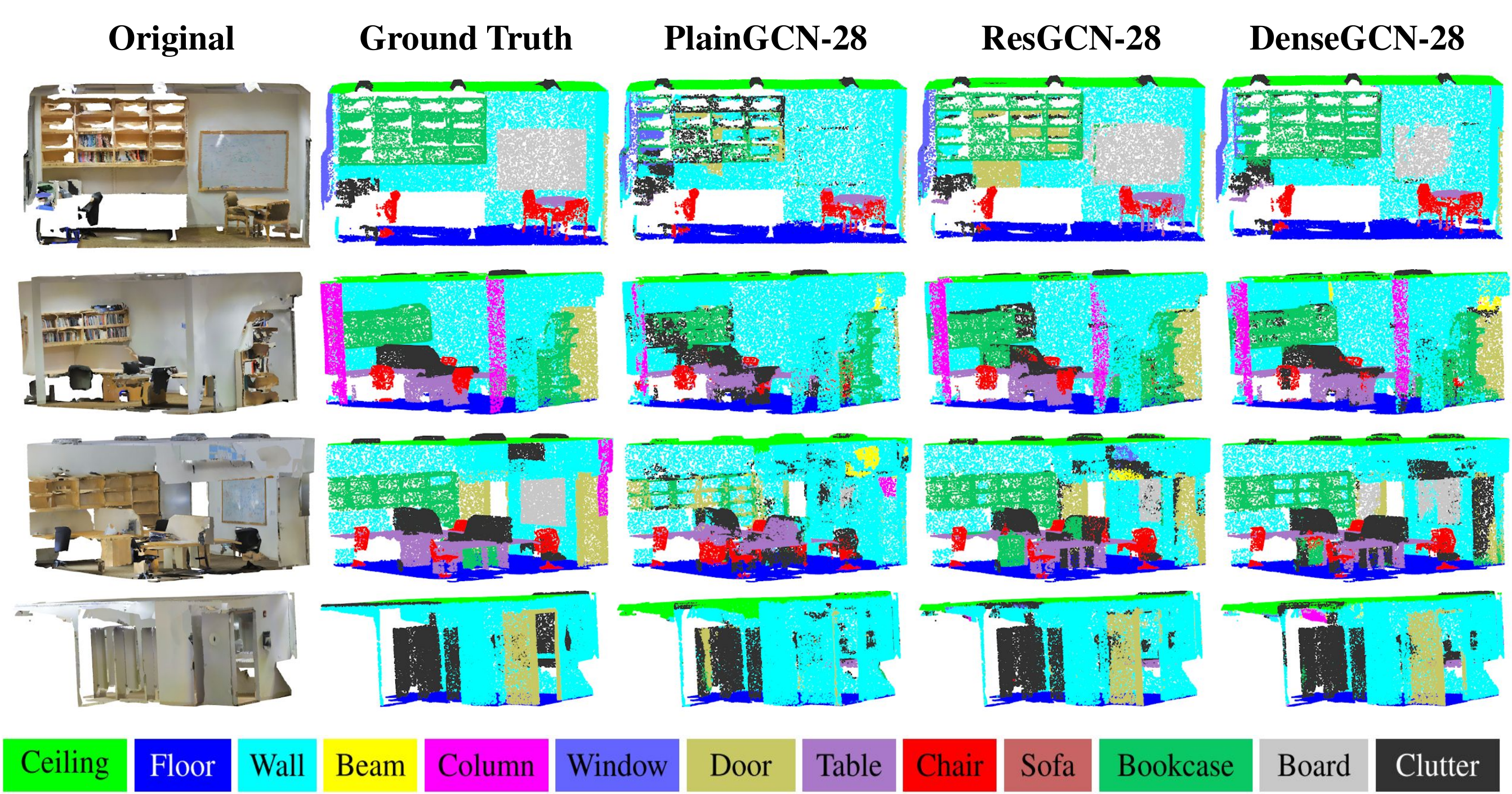}
    \caption{\textbf{Qualitative Results on S3DIS Semantic Segmentation}. We show here the effect of adding residual and dense graph connections to deep GCNs. \emph{PlainGCN-28}, \emph{ResGCN-28}, and \emph{DenseGCN-28} are identical except for the presence of residual graph connections in \emph{ResGCN-28} and dense graph connections in \emph{DenseGCN-28}. We note how both residual and dense graph connections have a substantial effect on hard classes like board, bookcase, and sofa. These are lost in the results of \emph{PlainGCN-28}.}
    \label{fig:qualitative}
\end{figure*}

\mysection{Qualitative Results}
\figLabel \ref{fig:qualitative} shows qualitative results on area 5 of S3DIS \cite{2017arXiv170201105A}. As expected from the results in \tblLabel \ref{tbl:ablation_area5}, our \emph{ResGCN-28} and \emph{DenseGCN-28} perform particularly well on difficult classes such as board, beam, bookcase and door. Rows 1-4 clearly show how \emph{ResGCN-28} and \emph{DenseGCN-28} are able to segment the board, beam, bookcase and door respectively, while \emph{PlainGCN-28} completely fails. Please refer to Appendices \ref{appendix:qualitative_results}, \ref{appendix:runtime_overhead} and \ref{appendix:dgcnn_all_classes} for more qualitative results and other ablation studies.

\begin{table*}[!htb]
\centering
\footnotesize 
\setlength{\tabcolsep}{4.2pt} 
\begin{tabular}{lcc|ccccccccccccc}
\toprule
\textbf{Method}  & \textbf{OA} & \textbf{mIOU}    & \textbf{ceiling} & \textbf{floor}   & \textbf{wall}    & \textbf{beam}    & \textbf{column}  & \textbf{window}  & \textbf{door}    & \textbf{table}   & \textbf{chair}   & \textbf{sofa}    & \textbf{bookcase} & \textbf{board}   & \textbf{clutter} \\
\midrule
PointNet \cite{pc_qi2017pointnet}& 78.5    & 47.6    & 88.0    & 88.7    & 69.3    & 42.4    & 23.1    & 47.5    & 51.6    & 54.1    & 42.0    & 9.6     & 38.2    & 29.4    & 35.2    \\
MS+CU \cite{3dsemseg_ICCVW17}   & 79.2    & 47.8    & 88.6    & \textbf{95.8} & 67.3    & 36.9    & 24.9    & 48.6    & 52.3    & 51.9    & 45.1    & 10.6    & 36.8    & 24.7    & 37.5    \\
G+RCU \cite{3dsemseg_ICCVW17}    & 81.1    & 49.7    & 90.3    & 92.1    & 67.9    & \textbf{44.7} & 24.2    & 52.3    & 51.2    & 58.1    & 47.4    & 6.9     & 39.0    & 30.0    & 41.9    \\
PointNet++ \cite {pc_qi2017pointnet++} &-  &53.2 &90.2 &91.7 &73.1 &42.7 &21.2 &49.7 &42.3 &62.7 &59.0 &19.6 &45.8 &48.2 &45.6\\
3DRNN+CF \cite{pc_ye20183d}& \textbf{86.9} & 56.3    & 92.9    & 93.8    & 73.1    & 42.5    & 25.9    & 47.6    & 59.2    & 60.4    & \textbf{66.7} & 24.8    & \textbf{57.0} & 36.7    & 51.6    \\
\midrule
DGCNN \cite{wang2018dynamic}  & 84.1    & 56.1    & - & - & - & - & - & - & - & - & - & - & - & - & - \\
\textbf{ResGCN-28 (\textit{Ours})} & 85.9    & \textbf{60.0} & \textbf{93.1} & 95.3    & \textbf{78.2} & 33.9    & \textbf{37.4} & \textbf{56.1} & \textbf{68.2} & \textbf{64.9} & 61.0  & \textbf{34.6}    & 51.5    & \textbf{51.1} & \textbf{54.4}\\
\bottomrule
\end{tabular}
\vspace{3pt}
\caption{\textbf{Comparison of \emph{ResGCN-28} with state-of-the-art on S3DIS Semantic Segmentation}. We report average per-class results across all areas for our reference model \emph{ResGCN-28}, which has 28 GCN layers, residual graph connections, and dilated graph convolutions, and state-of-the-art baselines. \emph{ResGCN-28} outperforms state-of-the-art by almost $4\%$. It also outperforms all baselines in $9$ out of $13$ classes. The metrics shown are overall point accuracy (OA) and mean IoU (mIoU). '-' denotes not reported and \textbf{bold} denotes best performance.}
\label{tbl:ours_vs_baselines}
\end{table*}

\mysection{Comparison to state-of-the-art} 
Finally, we compare our reference network (\emph{ResGCN-28}), which incorporates the ideas put forward in the methodology, to several state-of-the-art baselines in \tblLabel \ref{tbl:ours_vs_baselines}. The results clearly show the effectiveness of deeper models with residual graph connections and dilated graph convolutions. \emph{ResGCN-28} outperforms DGCNN \cite{wang2018dynamic} by 3.9\% (absolute) in mean IoU, even though DGCNN has the same fusion and MLP prediction blocks as \emph{ResGCN-28} but with a shallower \emph{PlainGCN} backbone block.  
Furthermore, we outperform all baselines in 9 out of 13 classes. We perform particularly well in the difficult object classes such as board, where we achieve 51.1\%, and sofa, where we improve state-of-the-art by about 10\%.

This significant performance improvement on the difficult classes is probably due to the increased network capacity, which allows the network to learn subtle details necessary to distinguish between a board and a wall for example. The first row in \figLabel \ref{fig:qualitative} is a representative example for this occurrence. 
Our performance gains are solely due to our innovation in the network architecture, since we use the same hyper-parameters and even learning rate schedule as the baseline DGCNN \cite{wang2018dynamic} and only decrease the number of nearest neighbors from $20$ to $16$ and the batch size from $24$ to $16$ due to memory constraints. We outperform state-of-the art methods by a significant margin and expect further improvement from tweaking the hyper-parameters, especially the learning schedule.

\section{Conclusion and Future Work}
\label{sec:conclusion}

In this work, we investigate how to bring proven useful concepts (residual connections, dense connections and dilated convolutions) from CNNs to GCNs and answer the question: how can GCNs be made deeper? Extensive experiments show that by adding skip connections to GCNs, we can alleviate the difficulty of training, which is the primary problem impeding GCNs to go deeper. Moreover, dilated graph convolutions help to gain a larger receptive field without loss of resolution. Even with a small amount of nearest neighbors, deep GCNs can achieve high performance on point cloud semantic segmentation. \emph{ResGCN-56} performs very well on this task, although it uses only $8$ nearest neighbors compared to $16$ for \emph{ResGCN-28}. We were also able to train \emph{ResGCN-151} for 80 epochs; the network converged very well and achieved similar results as \emph{ResGCN-28} and \emph{ResGCN-56} but with only 3 nearest neighbors. Due to computational constraints, we were unable to investigate such deep architectures in detail and leave it for future work. 

Our results show that after solving the vanishing gradient problem plaguing deep GCNs, we can either make GCNs deeper or wider (\eg \emph{ResGCN-28W}) to get better performance. We expect GCNs to become a powerful tool for processing non-Euclidean data in computer vision, natural language processing, and data mining. We show successful cases for adapting concepts from CNNs to GCNs. In the future, it will be worthwhile to explore how to transfer other operators, \eg deformable convolutions \cite{dai2017deformable}, other architectures, \eg feature pyramid architectures \cite{zhao2017pyramid}, \etc. It will also be interesting to study different distance measures to compute dilated $k$-NN, constructing graphs with different $k$ at each layer, better dilation rate schedules \cite{chen2017rethinking, wang2018understanding} for GCNs, and combining residual and dense connections. 

We also point out that, for the specific task of point cloud semantic segmentation, the common approach of processing the data in $1m\times 1m$ columns is sub-optimal for graph representation. A more suitable sampling approach should lead to further performance gains on this task.

\mysection{Acknowledgments} The authors thank Adel Bibi and Guocheng Qian for their help with the project. This work was supported by the King Abdullah University of Science and Technology (KAUST) Office of Sponsored Research through the Visual Computing Center (VCC) funding.

{\small
\bibliographystyle{ieee}
\bibliography{references}
}

\clearpage
\appendix

\section{Deep GCN Variants} \label{appendix:gcn_variants}
In our experiments in the paper, we work with a GCN based on EdgeConv \cite{wang2018dynamic} to show how very deep GCNs can be trained. However, it is straightforward to build other deep GCNs with the same concepts we proposed (\eg \emph{residual/dense graph connections}, \emph{dilated graph convolutions}). To show that these concepts are universal operators and can be used for general GCNs, we perform additional experiments. In particular, we build ResGCNs based on GraphSAGE \cite{hamilton2017inductive}, Graph Isomorphism Network (GIN) \cite{xu2018powerful} and \emph{MRGCN} (Max-Relative GCN) which is a new GCN operation we proposed. In practice, we find that EdgeConv learns a better representation than the other implementations. However, it is less memory and computation efficient. Therefore, we propose a simple GCN combining the advantages of them all. 

All of the ResGCNs have the same components (\eg dynamic $k-NN$, residual connections, stochastic dilation) and parameters (\eg \#NNs, \#filters and \#layers) as \emph{ResGCN-28} in Table \textbf{Ablation Study} of the paper except for the internal GCN operations. To simplify, we refer to these models as \emph{ResEdgeConv}, \emph{ResGraphSAGE}, \emph{ResGIN} and \emph{NewResGCN} respectively. Note that \emph{ResEdgeConv} is an alias for \emph{ResGCN} in our paper. We refer to it as \emph{ResEdgeConv} to distinguish it from the other GCN operations.

\mysection{ResEdgeConv} Instead of aggregating neighborhood features directly, EdgeConv \cite{wang2018dynamic} proposes to first get local neighborhood information for each neighbor by subtracting the feature of the central vertex from its own feature. In order to train deeper GCNs, we add \emph{residual/dense graph connections} and \emph{dilated graph convolutions} to EdgeConv:
\begin{equation}
\begin{split}
\mathbf{h}^{res}_{v_{l+1}} &=\textit{max}\left(\{\textit{mlp}(\textit{concat}(\mathbf{h}_{v_{l}}, \mathbf{h}_{u_{l}}-\mathbf{h}_{v_{l}}))|u_{l}\in \mathcal{N}^{(d)}(v_{l})\}\right), \\
\mathbf{h}_{v_{l+1}} &= \mathbf{h}^{res}_{v_{l+1}} + \mathbf{h}_{v_{l}}.
\end{split}
\end{equation}

\mysection{ResGraphSAGE} GraphSAGE \cite{hamilton2017inductive} proposes different types of aggregator functions including a \emph{Mean aggregator}, \emph{LSTM aggregator} and \emph{Pooling aggregator}. Their experiments show that the \emph{Pooling aggregator} outperforms the others. We adapt GraphSAGE with the max-pooling aggregator to obtain \emph{ResGraphSAGE}:

\begin{equation} \label{eq:sage}
\begin{split}
\mathbf{h}^{res}_{\mathcal{N}^{(d)}(v_{l})} &=\textit{max}\left(\{\textit{mlp}(\mathbf{h}_{u_{l}})|u_{l}\in \mathcal{N}^{(d)}(v_{l})\}\right), \\
\mathbf{h}^{res}_{v_{l+1}} &=\textit{mlp}\left(\textit{concat}\bigg(\mathbf{h}_{v_{l}}, \mathbf{h}^{res}_{\mathcal{N}^{(d)}(v_{l})}\bigg)\right), \\
\mathbf{h}_{v_{l+1}} &= \mathbf{h}^{res}_{v_{l+1}} + \mathbf{h}_{v_{l}},
\end{split}
\end{equation}

In the original GraphSAGE paper, the vertex features are normalized after aggregation. We implement two variants, one without normalization (see  \eqLabel \eqref{eq:sage}), the other one with normalization $\mathbf{h}^{res}_{v_{l+1}} = \mathbf{h}^{res}_{v_{l+1}} / \left\Vert \mathbf{h}^{res}_{v_{l+1}} \right\Vert_{2}$.

\mysection{ResGIN} The main difference between GIN \cite{xu2018powerful} and other GCNs is that an $\epsilon$ is learned at each GCN layer to give the central vertex and aggregated neighborhood features different weights. Hence \emph{ResGIN} is formulated as follows:
\begin{equation}
\begin{split}
\mathbf{h}^{res}_{v_{l+1}} &=\textit{mlp}\left((1+\epsilon)\cdot \mathbf{h}_{v_{l}} + \textit{sum}(\{\mathbf{h}_{u_{l}}|u_{l}\in \mathcal{N}^{(d)}(v_{l})\})\right), \\
\mathbf{h}_{v_{l+1}} &= \mathbf{h}^{res}_{v_{l+1}} + \mathbf{h}_{v_{l}}.
\end{split}
\end{equation}

\begin{table*}[!ht]
\centering
\small
\setlength{\tabcolsep}{7pt} 
\begin{tabular}{c|lcc|cccccccc}
\toprule
 \textbf{\rot{Model}} & \textbf{\rot{mIoU}} & \textbf{\rot{$\Delta$mIoU}} & \textbf{\rot{dynamic}} & \textbf{\rot{connection}} & \textbf{\rot{dilation}} & \textbf{\rot{stochastic}} & \textbf{\rot{\# NNs}} & \textbf{\rot{\# filters}} & \textbf{\rot{\# layers}} \\
\midrule
\emph{ResEdgeConv-28} & \textbf{52.49} & 0.00 & \checkmark  & \res & \checkmark & \checkmark & 16 & 64 & 28 \\
\midrule
\emph{PlainGCN-28} & \textbf{40.31} & -12.18 & \checkmark  & \nc &            &            & 16 & 64 & 28 \\
\midrule
\emph{ResGraphSAGE-28} & \textbf{49.20} & -3.29 & \checkmark  & \res & \checkmark & \checkmark & 16 & 64 & 28 \\
\midrule
\emph{ResGraphSAGE-N-28} & \textbf{49.02} & -3.47 & \checkmark  & \res & \checkmark & \checkmark & 16 & 64 & 28 \\
\midrule
\emph{ResGIN-$\epsilon$-28} & \textbf{42.81} & -9.68 & \checkmark  & \res & \checkmark & \checkmark & 16 & 64 & 28 \\
 \midrule
\emph{ResMRGCN-28} & \textbf{51.17} & -1.32 & \checkmark  & \res & \checkmark & \checkmark & 16 & 64 & 28 \\
\bottomrule
\end{tabular}
\vspace{3pt}
\caption{\textbf{Comparisons of Deep GCNs variants on area 5 of S3DIS}. We compare our different types of ResGCN (\emph{ResEdgeConv}, \emph{ResGraphSAGE},  \emph{ResGIN} and \emph{ResMRGCN}) with 28 layers. \emph{Residual graph connections} and \emph{Dilated graph convolutions} are added to all the GCN variants. All models were trained with the same hyper-parameters for 100 epochs on all areas except for area 5 which is used for evaluation. We denote residual with the \res~ symbols.}
\label{tbl:GCNs_variants}
\end{table*}

\mysection{ResMRGCN} We find that first using a max aggregator to aggregate neighborhood relative features $(\mathbf{h}_{u_{l}} - \mathbf{h}_{v_{l}}) ,~ u_{l}\in \mathcal{N}(v_{l})$ is more efficient than aggregating raw neighborhood features $\mathbf{h}_{v_{l}} ,~ u_{l}\in \mathcal{N}(v_{l})$ or aggregating features after non-linear transforms. We refer to this simple GCN as \emph{MRGCN} (Max-Relative GCN). The residual version of \emph{MRGCN} is as such:
\begin{equation}
\begin{split}
\mathbf{h}^{res}_{\mathcal{N}^{(d)}(v_{l})} &=\textit{max}\left(\{\mathbf{h}_{u_{l}}-\mathbf{h}_{v_{l}}|u_{l}\in \mathcal{N}^{(d)}(v_{l})\}\right), \\
\mathbf{h}^{res}_{v_{l+1}} &=\textit{mlp}\left(\textit{concat}\bigg(\mathbf{h}_{v_{l}}, \mathbf{h}^{res}_{\mathcal{N}^{(d)}(v_{l})}\bigg)\right), \\
\mathbf{h}_{v_{l+1}} &= \mathbf{h}^{res}_{v_{l+1}} + \mathbf{h}_{v_{l}}.
\end{split}
\end{equation}
Where $\mathbf{h}_{v_{l+1}}$ and $\mathbf{h}_{v_{l}}$ are the hidden state of vertex $v$ at $l+1$; $\mathbf{h}^{res}_{v_{l+1}}$ is the hidden state of the residual graph. All the \emph{mlp} (multilayer perceptron) functions use a ReLU as activation function; all the \emph{max} and \emph{sum} functions above are vertex-wise feature operators; \emph{concat} functions concatenate features of two vertices into one feature vector. $\mathcal{N}^{(d)}(v_{l})$ denotes the neighborhood of vertex $v_l$ obtained from \emph{Dilated $k$-NN}.

\section{Results for Deep GCN Variants}\label{appendix:results_gcn_variants}
\tblLabel \ref{tbl:GCNs_variants} shows a comparison of different deep residual GCNs variants on the task of semantic segmentation; we report the mIOU for area 5 of S3DIS. All deep GCN variants are 28 layers deep and we denote them as \emph{ResEdgeConv-28}, \emph{ResGraphSAGE-28}, \emph{ResGraphSAGE-N-28}, \emph{ResGIN-$\epsilon$-28} and \emph{ResMRGCN-28}; \emph{ResGraphSAGE-28} is GraphSAGE without normalization, \emph{ResGraphSAGE-N-28} is the version with normalization. The results clearly show that different deep GCN variants with \emph{residual graph connections} and \emph{dilated graph convolutions} converge better than the \emph{PlainGCN}. \emph{ResMRGCN-28} achieves almost the same performance as \emph{ResEdgeConv-28} while only using half of the GPU memory. \emph{ResGraphSAGE-28} and \emph{ResGraphSAGE-N-28} are slightly worse than \emph{ResEdgeConv-28} and \emph{ResMRGCN-28}. The results also show that using normalization for \emph{ResGraphSAGE} is not essential. Interestingly, we find that \emph{ResGIN-$\epsilon$-28} converges well during the training phase and has a high training accuracy. However, it fails to generalize to the test set. This phenomenon is also observed in the original paper \cite{xu2018powerful} in which they find setting $\epsilon$ to $0$ can get the best performance. Therefore, we can draw the conclusion that the concepts we proposed (\eg \emph{residual/dense graph connections} and \emph{dilated graph convolutions}) generalize well to different types of GCNs and enable training very deep GCNs.

\section{Qualitative Results for the Ablation Study} \label{appendix:qualitative_results}
\begin{figure} [!htb]
    \centering
    \includegraphics[trim = 2mm 0mm 0mm 0mm, clip, width=\columnwidth]{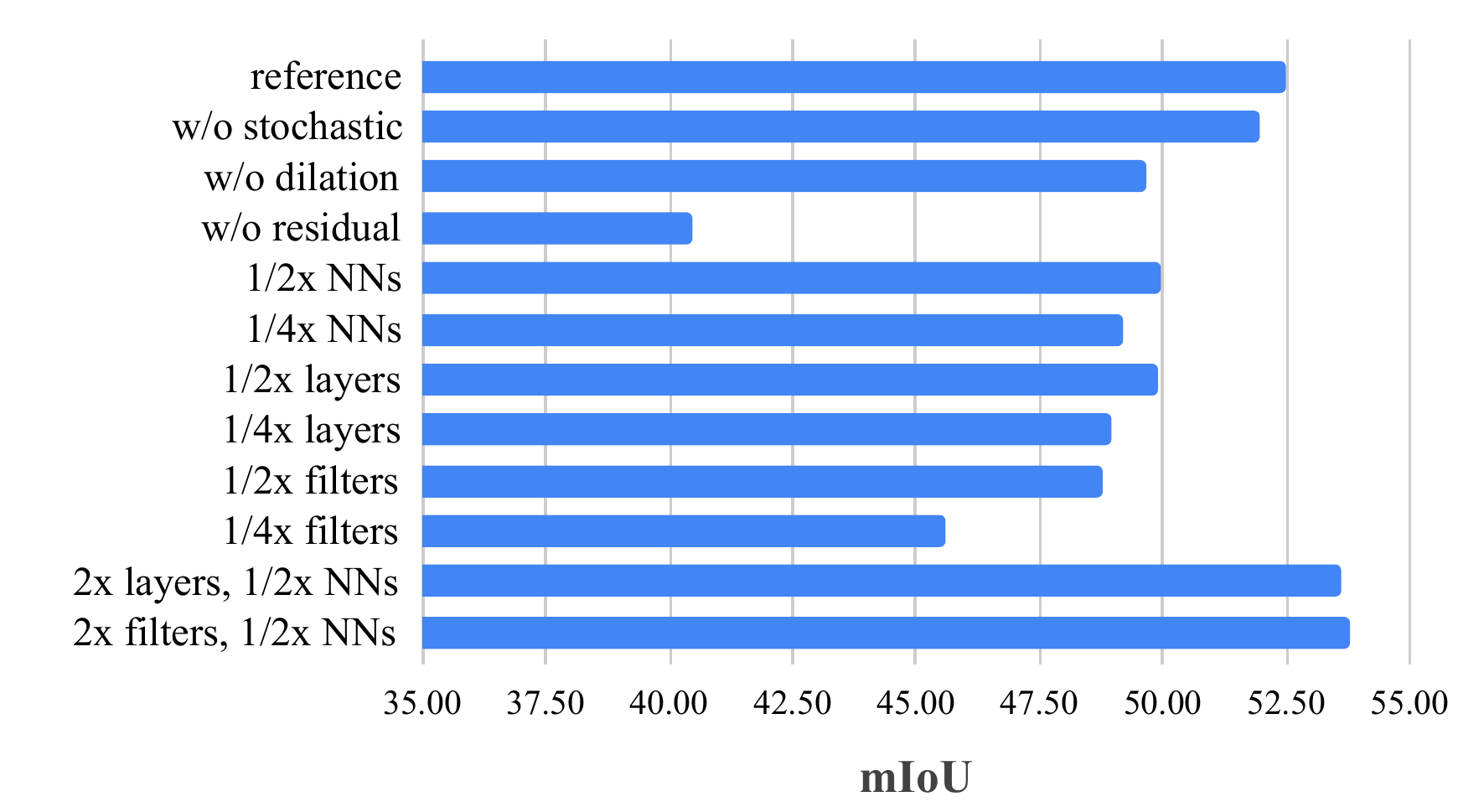}
    \caption{\textbf{Ablation study on area 5 of S3DIS}. We compare our reference network (\emph{ResGCN-28}) with 28 layers, \emph{residual graph connections} and \emph{dilated graph convolutions} to several ablated variants. All models were trained for 100 epochs on all areas except for area 5 with the same hyper-parameters.}
    \label{fig:ablation}
\end{figure}

\begin{figure*}[!h]
    \centering
    \includegraphics[height=\textheight]{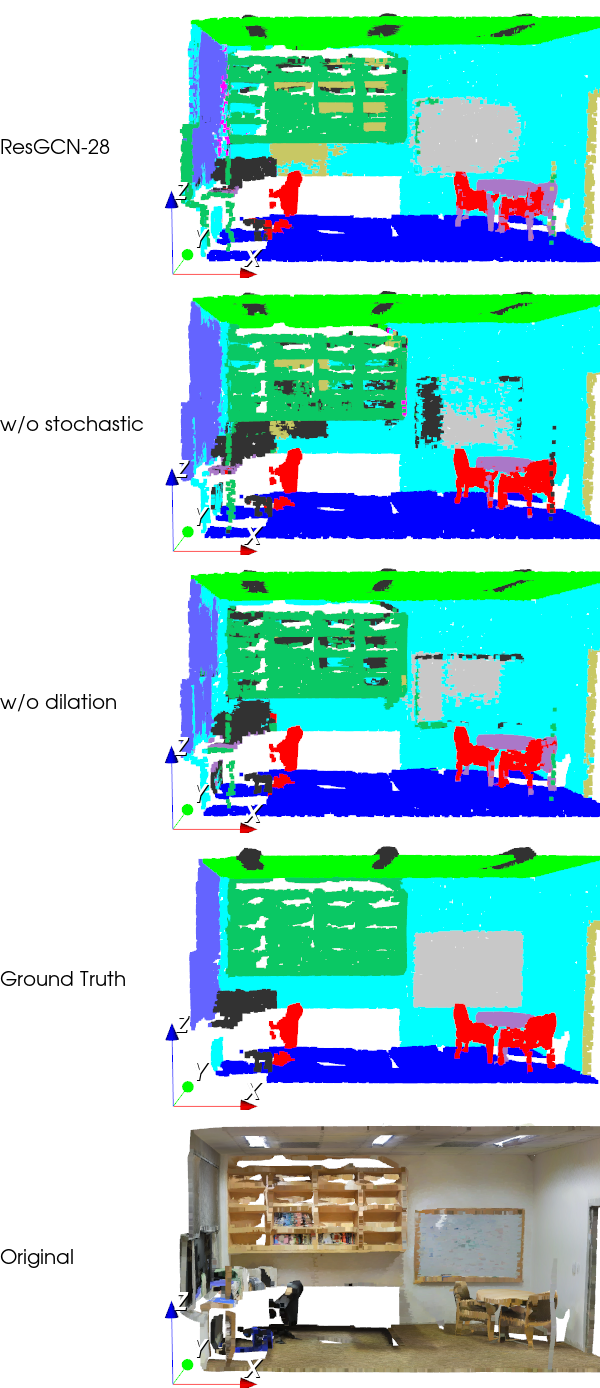}
    \caption{\textbf{Qualitative Results for S3DIS Semantic Segmentation}. We show the importance of stochastic dilated convolutions.}
    \label{fig:qualitative_dilation}
\end{figure*}

\begin{figure*}[!h]
    \centering
    \includegraphics[height=\textheight]{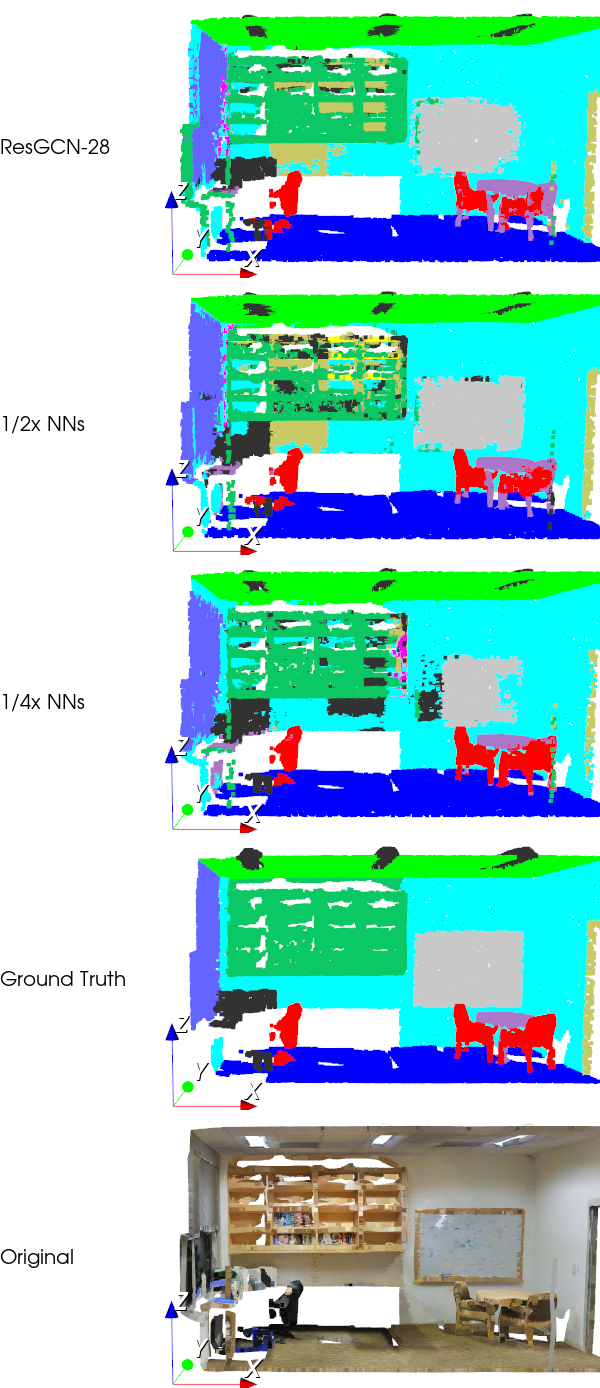}
    \caption{\textbf{Qualitative Results for S3DIS Semantic Segmentation}. We show the importance of the number of nearest neighbors used in the convolutions.}
    \label{fig:qualitative_nearest_neighbors}
\end{figure*}

\begin{figure*}[!h]
    \centering
    \includegraphics[height=\textheight]{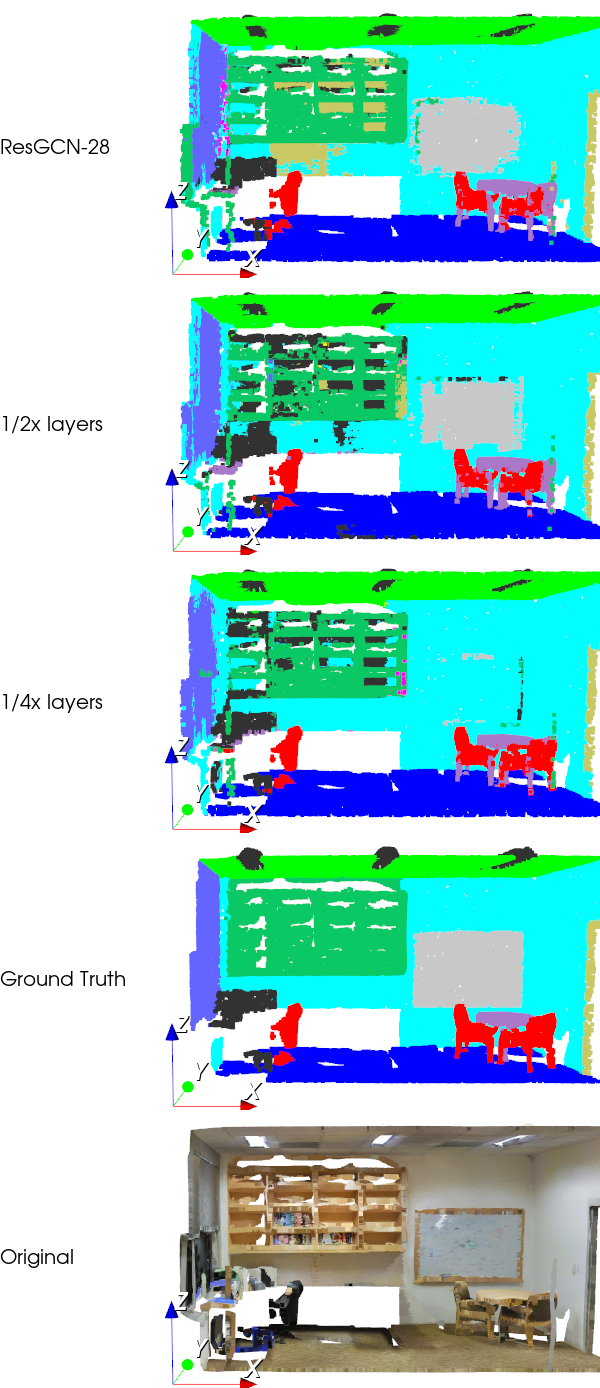}
    \caption{\textbf{Qualitative Results for S3DIS Semantic Segmentation}. We show the importance of network depth (number of layers).}
    \label{fig:qualitative_layers}
\end{figure*}

\begin{figure*}[!h]
    \centering
    \includegraphics[height=\textheight]{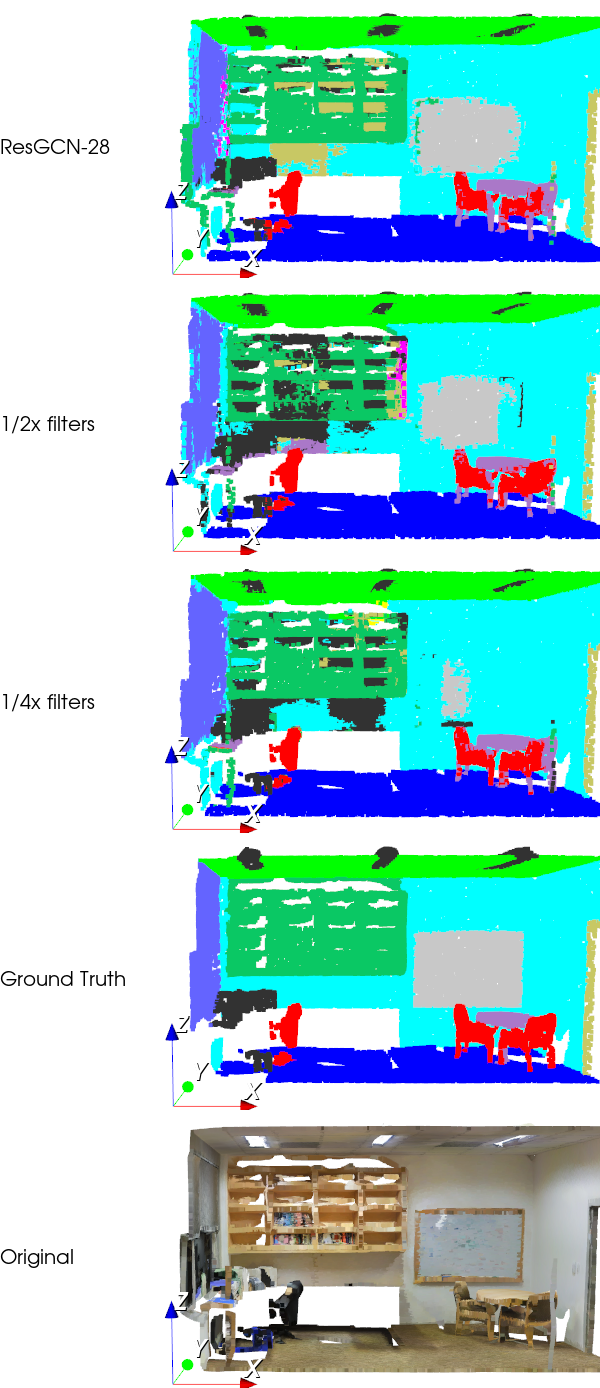}
    \caption{\textbf{Qualitative Results for S3DIS Semantic Segmentation}. We show the importance of network width (number of filters per layer).}
    \label{fig:qualitative_filters}
\end{figure*}

\begin{figure*}[!h]
    \centering
    \includegraphics[height=\textheight]{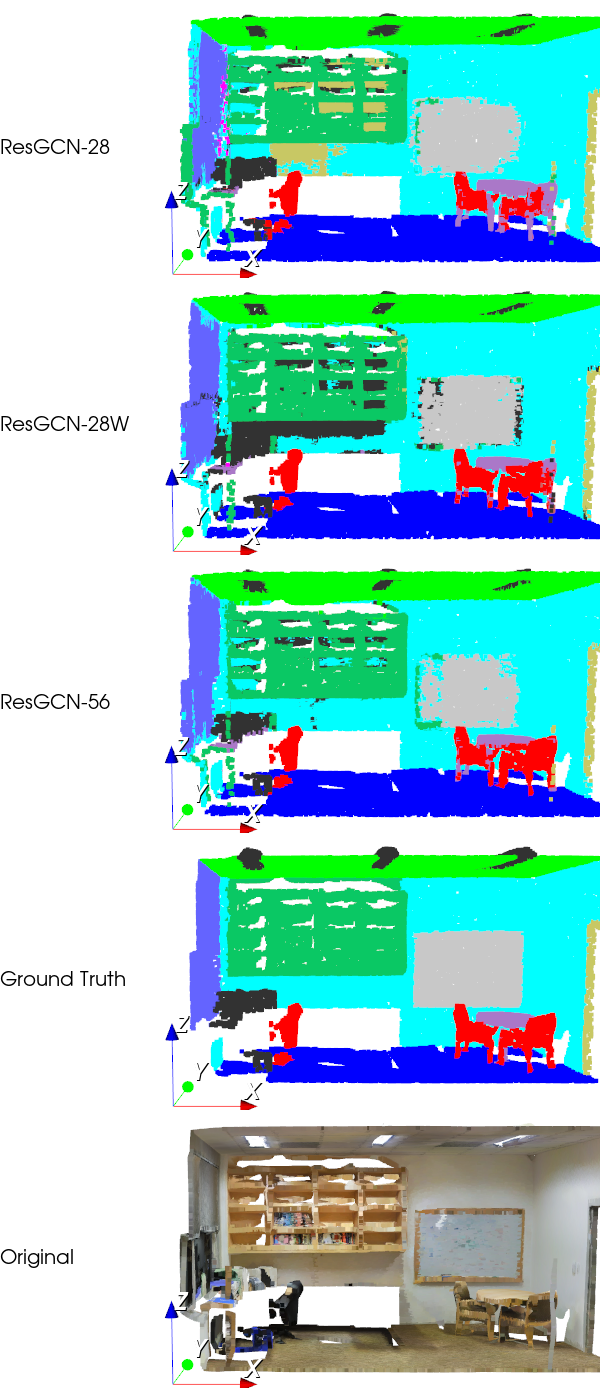}
    \caption{\textbf{Qualitative Results for S3DIS Semantic Segmentation}. We show the benefit of a wider and deeper network even with only half the number of nearest neighbors.}
    \label{fig:qualitative_56_28W}
\end{figure*}

We summarize the most important insights of the ablation study in \figLabel \ref{fig:ablation}. Figures \ref{fig:qualitative_dilation}, \ref{fig:qualitative_nearest_neighbors}, \ref{fig:qualitative_layers}, \ref{fig:qualitative_filters}, \ref{fig:qualitative_56_28W} show qualitative results for the ablation study presented in the paper.

\section{Run-time Overhead of Dynamic k-NN} \label{appendix:runtime_overhead}
We conduct a run-time experiment comparing the inference time of the reference model (28 layers, $k$=16) with dynamic k-NN and fixed k-NN. The inference time with fixed k-NN is 45.63ms. Computing the dynamic k-NN increases the inference time by 150.88ms. It is possible to reduce computation by updating the k-NN less frequently (\eg computing the dynamic k-NN every 3 layers).

\section{Comparison with DGCNN over All Classes} \label{appendix:dgcnn_all_classes}
To showcase the consistent improvement of our framework over the baseline DGCNN \cite{wang2018dynamic}, we reproduce the results of DGCNN\footnote{The results over all classes were not provided in the original DGCNN paper} in \tblLabel \ref{tbl:ours_vs_dgcnn} and find our method outperforms DGCNN in all classes.

\begin{table}[h]
\small
\centering
\setlength{\tabcolsep}{6pt} 
\begin{tabular}{l|cc}
\toprule
\textbf{Class}  & \textbf{DGCNN \cite{wang2018dynamic}} & \textbf{ResGCN-28 (\textit{Ours})} \\
\midrule
{ceiling} & 92.7 & \textbf{93.1} \\
{floor}   & 93.6 & \textbf{95.3} \\
{wall}    & 77.5 & \textbf{78.2} \\
{beam}    & 32.0 & \textbf{33.9} \\
{column}  & 36.3 & \textbf{37.4} \\
{window}  & 52.5 & \textbf{56.1} \\
{door}    & 63.7 & \textbf{68.2} \\
{table}   & 61.1 & \textbf{64.9} \\
{chair}   & 60.2 & \textbf{61.0} \\
{sofa}    & 20.5 & \textbf{34.6} \\
{bookcase}& 47.7 & \textbf{51.5} \\
{board}   & 42.7 & \textbf{51.1} \\
{clutter} & 51.5 & \textbf{54.4} \\
\midrule
\textbf{mIOU}    & 56.3 & \textbf{60.0} \\
\bottomrule
\end{tabular}
\vspace{2pt}
\caption{\textbf{Comparison of \emph{ResGCN-28} with DGCNN}. Average per-class results across all areas for our reference network with 28 layers, \emph{residual graph connections} and \emph{dilated graph convolutions} compared to DGCNN baseline. \emph{ResGCN-28} outperforms DGCNN across all the classes. Metric shown is IoU.}
\label{tbl:ours_vs_dgcnn}
\end{table}

\end{document}